\documentclass[conference]{IEEEtran}
\IEEEoverridecommandlockouts
\usepackage{cite}
\usepackage{listings}
\usepackage{xcolor}
\usepackage{tcolorbox}
\tcbuselibrary{listings, skins, breakable}
\usepackage{fontawesome5}
\usepackage{amsmath,amssymb,amsfonts}
\usepackage{orcidlink}
\usepackage{graphicx}
\usepackage{textcomp}
\usepackage{booktabs}
\usepackage{multirow}
\usepackage[font=small,skip=2pt]{caption}
\usepackage{hyperref}
\usepackage{url}
\usepackage{colortbl}
\usepackage{nicefrac}
\definecolor{bestcell}{HTML}{E8F5E9}
\usepackage{microtype}
\usepackage{xcolor}
\usepackage{algorithm}
\usepackage{algpseudocode}
\usepackage{tikz}
\usepackage{comment}
\usepackage{amssymb}
\usepackage{braket}
\usepackage{pifont}

\def\BibTeX{{\rm B\kern-.05em{\sc i\kern-.025em b}\kern-.08em
    T\kern-.1667em\lower.7ex\hbox{E}\kern-.125emX}}

\definecolor{codebg}{HTML}{F6F8FA}
\definecolor{codeborder}{HTML}{D0D7DE}
\definecolor{keywordcolor}{HTML}{CF222E}
\definecolor{stringcolor}{HTML}{0A3069}
\definecolor{commentcolor}{HTML}{6E7781}
\definecolor{funccolor}{HTML}{8250DF}
\definecolor{promptsys}{HTML}{6639BA}
\definecolor{promptuser}{HTML}{0969DA}
\definecolor{retaingreen}{HTML}{1A7F37}
\definecolor{rejectred}{HTML}{CF222E}
\definecolor{stagebg}{HTML}{F0F3F6}
\definecolor{promptasst}{HTML}{1A7F37}

\definecolor{keywordgreen}{HTML}{008000}
\definecolor{nameblue}{HTML}{0000FF}
\definecolor{textblack}{HTML}{111111}
\definecolor{commentgray}{HTML}{6E7781}
\lstdefinestyle{pythoncode}{
  language=Python,
  basicstyle=\ttfamily\small\color{textblack},
  backgroundcolor=\color{codebg},
  frame=single,
  rulecolor=\color{codeborder},
  framerule=0.4pt,
  framesep=8pt,
  xleftmargin=8pt,
  xrightmargin=8pt,
  breaklines=true,
  breakatwhitespace=false,
  prebreak=\mbox{\textcolor{textblack}{$\hookleftarrow$}},
  numbers=none,
  tabsize=4,
  showstringspaces=false,
  columns=fullflexible,
  keepspaces=true,
  keywordstyle=\color{keywordgreen}\bfseries,
  stringstyle=\color{nameblue},
  commentstyle=\color{commentgray}\itshape,
  identifierstyle=\color{textblack},
  emph={
    pennylane,qml,quantum_circuit,n_qubits,
    inputs,weights,wires,range,i
  },
  emphstyle=\color{nameblue},
  emph={[2]import,as,def,return,for,in},
  emphstyle={[2]\color{keywordgreen}\bfseries},
}
\lstdefinestyle{pennylanecode}{
  language=Python,
  basicstyle=\ttfamily\small,
  keywordstyle=\color{keywordcolor}\bfseries,
  stringstyle=\color{stringcolor},
  commentstyle=\color{commentcolor}\itshape,
  emph={qml, pennylane, AngleEmbedding, StronglyEntanglingLayers,
        PauliZ, expval, qnode, device, templates},
  emphstyle=\color{funccolor}\bfseries,
  backgroundcolor=\color{codebg},
  frame=single,
  rulecolor=\color{codeborder},
  framesep=6pt,
  xleftmargin=8pt,
  xrightmargin=8pt,
  breaklines=true,
  columns=fullflexible,
  keepspaces=true,
  breakindent=20pt,
  postbreak=\mbox{\textcolor{commentcolor}{$\hookrightarrow$}\space},
  lineskip=2pt,
  showstringspaces=false,
  tabsize=4,
  numbers=left,
  numberstyle=\tiny\color{commentcolor},
  numbersep=8pt,
  aboveskip=10pt,
  belowskip=10pt,
}

\newtcolorbox{systemprompt}[1][]{
colback=promptsys!5, colframe=promptsys!60,
coltitle=white, fonttitle=\bfseries\small\ttfamily,
title={\faRobot~SYSTEM}, arc=2pt, boxrule=0.6pt,
left=6pt, right=6pt, top=4pt, bottom=4pt,
breakable, #1
}

\newtcolorbox{assistantprompt}[1][]{
  colback=promptasst!5, colframe=promptasst!60,
  coltitle=white, fonttitle=\bfseries\small\ttfamily,
  title={\faComment~ASSISTANT}, arc=2pt, boxrule=0.6pt,
  left=6pt, right=6pt, top=4pt, bottom=4pt,
  breakable, #1
}

\newtcolorbox{userprompt}[1][]{
colback=promptuser!5, colframe=promptuser!60,
coltitle=white, fonttitle=\bfseries\small\ttfamily,
title={\faUser~USER}, arc=2pt, boxrule=0.6pt,
left=6pt, right=6pt, top=4pt, bottom=4pt,
breakable, #1
}

\newtcolorbox{stagebox}[2][]{
colback=stagebg, colframe=codeborder,
coltitle=black, fonttitle=\bfseries,
title={#2}, arc=3pt, boxrule=0.5pt,
left=8pt, right=8pt, top=6pt, bottom=6pt,
breakable, #1
}

\begin{document}
\bstctlcite{bstctl:nodash}

\title{PennySynth: RAG-Driven Data Synthesis for Automated Quantum Code Generation}

%\author{\IEEEauthorblockN{Anonymous Authors}}
\author{Minghao Shao\orcidlink{0009-0002-4467-6224}\textsuperscript{1,3}, Nouhaila Innan\orcidlink{0000-0002-1014-3457}\textsuperscript{1,2}, Hariharan Janardhanan\orcidlink{0009-0004-8930-2765}\textsuperscript{1,2}, Muhammad Kashif\orcidlink{0000-0003-2023-6371}\textsuperscript{1,2},\\ Alberto Marchisio\orcidlink{0000-0002-0689-4776}\textsuperscript{1,2}, and Muhammad Shafique\orcidlink{0000-0002-2607-8135}\textsuperscript{1,2}\\
 \IEEEauthorblockA{
 \textsuperscript{1}eBRAIN Lab, Division of Engineering, New York University Abu Dhabi (NYUAD), Abu Dhabi, UAE\\
 \textsuperscript{2}Center for Quantum and Topological Systems (CQTS), NYUAD Research Institute, NYUAD, Abu Dhabi, UAE\\
  \textsuperscript{3}Department of Computer Science and Engineering, NYU Tandon School of Engineering, New York, USA\\
 \{shao.minghao, nouhaila.innan, hj2342, muhammadkashif, alberto.marchisio, muhammad.shafique\}@nyu.edu\\
}}
\begin{comment}
\author{
  \IEEEauthorblockN{[Author],
    Nouhaila Innan\textsuperscript{1,2},
    Muhammad Shafique\textsuperscript{1,2}}
  \IEEEauthorblockA{
    \textsuperscript{1}eBRAIN Lab, Division of Engineering,
    New York University Abu Dhabi (NYUAD), Abu Dhabi, UAE\\
    \textsuperscript{2}Center for Quantum and Topological Systems (CQTS),
    NYUAD Research Institute, Abu Dhabi, UAE
  }
}
\end{comment}

\maketitle

%% ─────────────────────────────────────────────────────────────
\begin{abstract}
The growing complexity of quantum programming frameworks has exposed a critical limitation in existing large language model (LLM)-based code assistants: general-purpose models hallucinate PennyLane-specific gate names, misplace device configurations, and produce structurally invalid circuits when faced with specialized quantum coding challenges. We present \textbf{PennySynth}, a retrieval-augmented generation framework that addresses this gap by conditioning LLM inference on a curated knowledge base of 13,389 PennyLane instruction-code pairs, built via a three-stage extraction, verification, and deduplication pipeline over official PennyLane repositories, community GitHub sources, and QHack competition archives. PennySynth introduces a code-aware embedding strategy using \texttt{st-codesearch-distilroberta-base}, trained for natural-language-to-code retrieval, increasing average retrieval cosine similarity from 0.45 to 0.726 compared to a general-purpose baseline. Evaluated across 74 challenges spanning three years of the QHack competition (2022, 2023, 2024), PennySynth achieves \textbf{64\%}, \textbf{68\%}, and \textbf{52\%} pass@5 on QHack 2022, 2023, and 2024, respectively, improving over Claude Sonnet 4.6 without retrieval by \textbf{+28}, \textbf{+25}, and \textbf{+28} percentage points. We further introduce a quantum-adapted CodeBLEU metric that upweights \texttt{qml.*} token patterns and show that structural code similarity and functional correctness capture distinct aspects of quantum code quality. Controlled ablations reveal that code-aware embeddings are the primary driver of retrieval performance, while dataset expansion and source composition provide additional gains when retrieval quality is sufficiently precise.
\end{abstract}


%% ─────────────────────────────────────────────────────────────
\section{Introduction}

Quantum computing is progressing from theoretical foundations toward increasingly practical software and hardware ecosystems. Recent hardware advances have expanded the scale of available quantum processors, while software frameworks such as PennyLane~\cite{bergholm2018pennylane}, Qiskit \cite{javadi2024quantum}, and related quantum programming toolkits have made quantum and hybrid quantum-classical programming more accessible to researchers and developers. At the same time, quantum software development remains difficult because programs must correctly combine circuit construction, device configuration, differentiable execution, measurement design, and version-specific API behavior \cite{zaman2023qml}. These challenges create a growing need for intelligent coding tools that can assist developers in writing correct, efficient, and framework-compatible quantum programs.

Large language models (LLMs) have substantially improved general-purpose code generation~\cite{chen2021codex,lozhkov2024starcoder2}, and recent work has begun to extend these capabilities to quantum programming. For Qiskit, domain-specific models and tools, including IBM's Qiskit Code Assistant, have been developed using curated quantum code corpora~\cite{dupuis2024qiskit}, while Qiskit HumanEval~\cite{vishwakarma2024qiskithumaneval} provides a dedicated benchmark for executable Qiskit code generation. Beyond Qiskit, QuanBench~\cite{guo2025quanbench} evaluates LLM-based quantum code generation across a broader set of quantum programming tasks. For PennyLane, PennyLang~\cite{basit2025pennylang}, PennyCoder~\cite{basit2025pennycoder}, and QHackBench~\cite{basit2025qhackbench} establish datasets, models, and competition-grade evaluation settings for LLM-assisted PennyLane programming. The AWS Braket RAG system~\cite{kharkov2024braket} further shows that retrieval-augmented generation can support quantum program development. Despite this progress, generating correct quantum code remains challenging, particularly when outputs are evaluated through execution rather than textual similarity.

While recent efforts show that LLMs can support quantum code generation, reliable execution remains difficult, and two main bottlenecks continue to limit progress. First, quantum programming is a low-resource domain: PennyLane-specific code represents only a small fraction of the training data used by general-purpose LLMs, which limits their coverage of framework-specific operations, device declarations, templates, differentiable workflows, and measurement patterns. Second, PennyLane evolves rapidly, with API interfaces, gate naming conventions, device usage rules, and supported operations changing across releases. As a result, code samples that were valid under earlier versions may become outdated, increasing the risk of unsupported operations, incorrect device usage, missing imports, or structurally invalid circuits~\cite{basit2025qhackbench}. These limitations suggest that model scale alone is not sufficient; LLMs also need access to verified, up-to-date, and framework-specific code examples at inference time.

\begin{figure}[ht!]
  \centering
  \vspace{-4pt}
  \includegraphics[width=\columnwidth]{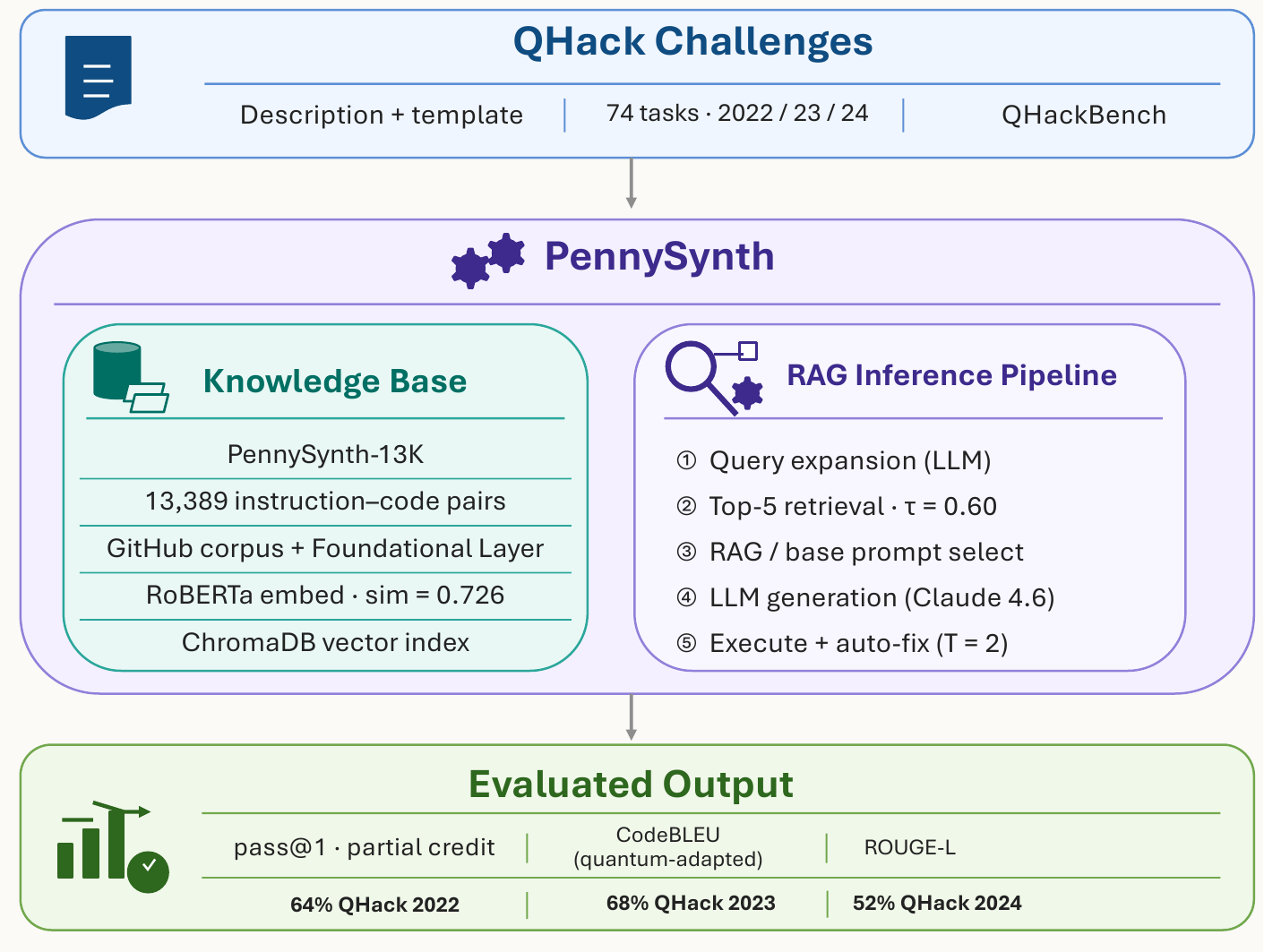}
  \caption{Overview of the PennySynth framework for automated PennyLane code generation. The system constructs a verified instruction-code knowledge base, retrieves task-relevant examples for each challenge, and uses retrieval-augmented generation with iterative execution-based repair to produce final solutions.}
  \label{fig:system}
  \vspace{-4pt}
\end{figure}

To address these limitations, we propose \textbf{PennySynth}, a data synthesis and retrieval framework for LLM-assisted PennyLane code generation. PennySynth operates in two stages. In the offline stage, it automatically extracts, verifies, and deduplicates PennyLane code from GitHub repositories, official documentation, and competition archives using AST-based function isolation, multi-layer quality verification, and MinHash-based deduplication. In the online stage, it retrieves task-relevant examples at inference time using a code-aware embedding model trained for natural-language-to-code retrieval. The resulting dataset, \textbf{PennySynth-13K}, contains 13,389 verified instruction-code pairs. Evaluated on QHack challenges, PennySynth achieves up to 68\% pass@5 on QHack 2023 with Claude Sonnet 4.6 as the base generator, corresponding to a +25 percentage point improvement over the same model without retrieval. Fig.~\ref{fig:system} illustrates the full pipeline.

Our main contributions are:
\begin{enumerate}
  \item \textbf{PennySynth}, an automated data synthesis framework for PennyLane quantum code that extracts, verifies, and deduplicates instruction-code pairs from heterogeneous sources, and \textbf{PennySynth-13K}, the resulting dataset of 13,389 AST-verified pairs released to support future research.
  \item \textbf{Code-aware RAG pipeline} with query expansion, relevance threshold, and selective context instructions, improving pass@5 over Claude Sonnet 4.6 without retrieval by up to +28.2 percentage points.
  \item \textbf{Quantum-adapted CodeBLEU}: upweighting \texttt{qml.*} tokens; first systematic CodeBLEU evaluation on generated PennyLane code vs.\ official QHack reference templates.
  \item \textbf{Systematic multi-model comparison} across seven state-of-the-art LLMs (Claude Sonnet 4.6, GPT-5.5, Gemini 2.5 Pro, Qwen3-235B-A22B, GLM-5.1, DeepSeek-V3, LLaMA 3.1-8B), with ablations isolating the effects of the embedding model, dataset size, and source composition.
\end{enumerate}

%% ─────────────────────────────────────────────────────────────
\section{Background and Related Work}

\subsection{LLMs for Quantum Code Generation}

LLM-based quantum code assistance has gained increasing attention across several quantum programming frameworks. For Qiskit, a domain-specific model based on the Granite architecture was fine-tuned on a curated corpus filtered to include post-2023 sources, reducing the risk of deprecated API usage~\cite{dupuis2024qiskit}. Qiskit HumanEval~\cite{vishwakarma2024qiskithumaneval} further provides more than 100 hand-curated quantum programming tasks across eight categories, ranging from basic circuit construction to more complex algorithm implementation. These efforts establish a foundation for standardized evaluation of LLM-generated Qiskit code.

For PennyLane, PennyLang~\cite{basit2025pennylang} provides 3,347 PennyLane-specific code samples collected from textbooks, documentation, and open-source repositories. When used within a RAG pipeline, PennyLang increased Qwen~7B's success rate from 8.7\% to 41.7\%, although stronger commercial models showed smaller gains from retrieval augmentation. PennyCoder~\cite{basit2025pennycoder} fine-tunes LLaMA 3.1-8B with LoRA for local deployment without external API dependencies, achieving 44.3\% accuracy on a 264-task internal benchmark and outperforming its RAG-only baseline at 40.2\%. However, its evaluation remains limited to in-domain tasks drawn from the PennyLang distribution. QHackBench~\cite{basit2025qhackbench} addresses this limitation by introducing a reproducible benchmark based on real QHack competition challenges and incorporating iterative refinement to correct generated solutions. Our work builds on this direction by introducing a larger verified PennyLane corpus and a code-aware retrieval pipeline, and evaluating multiple recent LLMs across QHack 2022, 2023, and 2024.

\subsection{Retrieval-Augmented Generation for Code}

Retrieval-Augmented Generation (RAG) conditions language model outputs on external examples or documents retrieved at inference time~\cite{lewis2020retrieval}. This approach reduces reliance on parametric memory alone and can provide models with task-specific context that may be absent or under represented in pretraining. RAG has also been extended through query expansion, where an LLM generates enriched search queries or pseudo-documents to improve retrieval quality~\cite{wang2023query2doc}, and through reranking methods that rescore retrieved candidates for finer-grained relevance estimation~\cite{nogueira2019passage}.

For code generation, retrieving semantically related code snippets or documentation fragments has been shown to improve generation accuracy compared with standalone LLMs~\cite{parvez2021retrieval}. In quantum computing, the AWS Braket RAG system~\cite{kharkov2024braket} demonstrated that retrieved documentation can support quantum program development. However, this system focuses on Braket-oriented workflows and does not target PennyLane competition-style challenges. More broadly, Graph RAG~\cite{edge2024local} organizes retrieved knowledge through graph structures, which can capture relations among entities and concepts. Such graph-based retrieval may be useful for quantum programming, where gates, templates, devices, measurements, and differentiation methods are strongly interconnected. In this work, we focus on a flat code-aware vector retrieval pipeline and leave PennyLane-specific graph retrieval as a future extension.

\begin{figure*}[t]
  \centering
  \includegraphics[width=\textwidth]{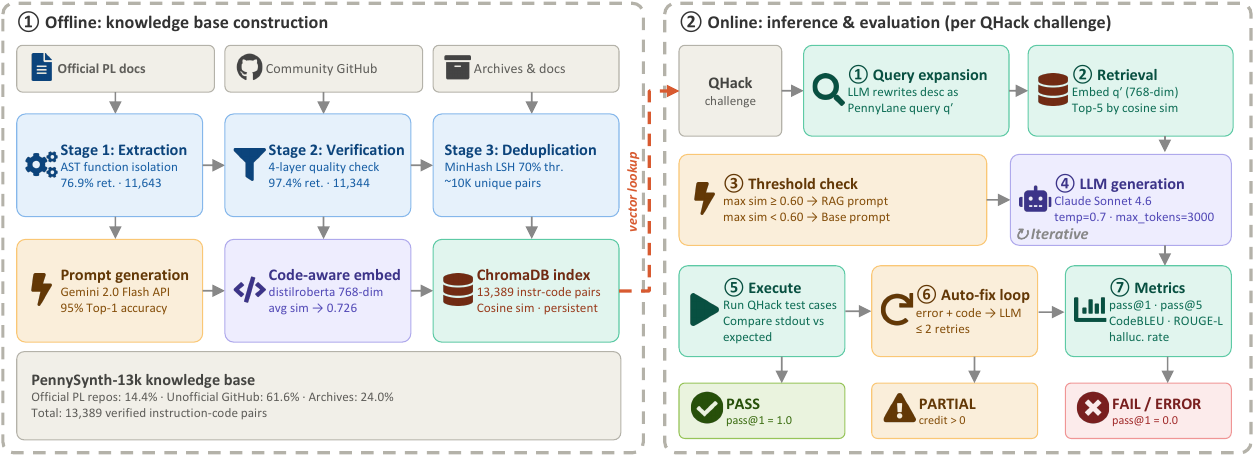}
  \caption{PennySynth full system architecture. \textbf{Offline:} three-stage knowledge base construction from heterogeneous PennyLane sources, followed by code-aware embedding into ChromaDB. \textbf{Online:} per-challenge inference via query expansion, top-5 retrieval, similarity threshold check, LLM generation, and iterative auto-fix feedback loop.}
  \label{fig:fullsystem}
\end{figure*}

\subsection{Evaluation Metrics}

BLEU~\cite{papineni2002bleu} measures n-gram precision with a brevity penalty and remains widely used in code generation evaluation. However, it is limited for code because it is sensitive to surface-level token overlap and does not account well for variable renaming, control-flow structure, or semantically equivalent implementations. ROUGE-L~\cite{lin2004rouge} measures the longest common subsequence between generated and reference outputs, providing an additional view of textual similarity.

CodeBLEU~\cite{ren2020codebleu} extends BLEU by combining token-level matching with structural and semantic code information. It includes token BLEU, weighted BLEU, AST matching, and dataflow matching, typically combined with equal weights. However, standard CodeBLEU does not explicitly distinguish framework-specific operations from generic tokens, even though API calls such as \texttt{qml.*} operations are central to the correctness of PennyLane programs. To better reflect PennyLane code structure, we adapt CodeBLEU in two ways: first, \texttt{qml.*} tokens are upweighted by 3$\times$ in the weighted BLEU component; second, the dataflow component is replaced with a quantum-oriented Jaccard match over extracted \texttt{qml.*} gate names, device declarations, and measurement return types. This provides a more domain-aware similarity measure for comparing generated PennyLane code with reference solutions.

%% ─────────────────────────────────────────────────────────────
\section{Methodology}
Fig.~\ref{fig:fullsystem} presents the overall PennySynth pipeline, which consists of two main stages: an \emph{offline} knowledge-base construction stage and an \emph{online} inference-and-repair stage. In the offline stage, PennySynth builds a verified repository of PennyLane instruction-code pairs collected from official sources, community repositories, and QHack archives. In the online stage, the system retrieves task-relevant examples for each challenge, adds them to the generation prompt when retrieval confidence is sufficiently high, and applies an execution-based repair loop when the generated code fails the test cases.

\subsection{PennySynth-13K Dataset}
\label{sec:dataset}

\subsubsection{Data Collection}

The quality of retrieval-augmented code generation depends on the coverage, correctness, and relevance of the retrieval corpus. To construct a knowledge base that captures both canonical PennyLane usage and realistic community coding patterns, we aggregate source code from three types of sources.

\begin{enumerate}
    \item \textbf{Official PennyLane sources:} We collect code from official PennyLane repositories and documentation examples, which provide trusted reference implementations of core PennyLane functionality.

    \item \textbf{Community GitHub repositories:} We gather a large community corpus from GitHub by searching for PennyLane-related keywords such as \texttt{pennylane} and \texttt{qml}, yielding more than 285,000 candidate records before filtering.

    \item \textbf{QHack competition archives:} We include public challenge descriptions and template code from the QHack 2022, 2023, and 2024 archives. These entries are valuable because they are aligned with the competition-style PennyLane programming tasks used in our evaluation. Since user-submitted solutions may appear in public repositories, we exclude solution-like files and files matching benchmark answer patterns from the knowledge base. This prevents direct answer retrieval during evaluation and ensures that performance gains come from task-relevant retrieved examples rather than memorized solutions.
\end{enumerate}

This source selection is designed to balance correctness, diversity, and benchmark relevance. Official sources provide trusted API usage, community repositories expand coverage of real-world PennyLane coding patterns, and QHack archives provide examples aligned with competition-style synthesis tasks.

\subsubsection{Dataset Profiling}
\label{sec:profiling}

Table~\ref{tab:dataset} summarizes the composition of PennySynth-13K across three source groups: official PennyLane repositories, unofficial GitHub repositories, and QHack/tutorial archives. The final dataset contains 13,389 instruction-code pairs. Unofficial GitHub repositories contribute the largest share, with 8,245 pairs, corresponding to 61.6\% of the corpus. Official PennyLane repositories contribute 1,934 pairs, corresponding to 14.4\%. These two GitHub-derived sources form a 10,179-pair corpus before archive augmentation. The remaining 3,210 pairs, corresponding to 24.0\%, come from QHack and tutorial archives.

\begin{table}[ht!]
\centering
\caption{PennySynth-13K Dataset Composition by Source}
\label{tab:dataset}
\setlength{\tabcolsep}{3pt}
\begin{tabular}{llcc}
\toprule
\textbf{Category} & \textbf{Source} & \textbf{Samples} & \textbf{\%} \\
\midrule
Official  & Official PennyLane repos    & 1,934  & 14.4\% \\
Community & Unofficial GitHub repos     & 8,245  & 61.6\% \\
Archives  & QHack \& Tutorial archives  & 3,210  & 24.0\% \\
\midrule
\multicolumn{2}{l}{\textbf{Total (after deduplication + archives)}}
  & \textbf{13,389} & \textbf{100\%} \\
\bottomrule
\end{tabular}
\end{table}
This composition reflects the role of each source group in the final corpus. Unofficial GitHub repositories provide broad coverage of real-world PennyLane coding patterns, official repositories provide trusted framework-level examples, and QHack/tutorial archives add benchmark-aligned prompts and template structures. This breakdown also motivates the dataset ablations reported later, where we compare the 10,179-pair GitHub-derived corpus against the full 13,389-pair PennySynth-13K corpus and further analyze the contribution of each source category.

\subsubsection{Three-Stage Dataset Construction Pipeline}

We transform raw source files into verified instruction-code pairs through a three-stage pipeline: extraction, verification, and deduplication. The pipeline is designed to preserve executable structure and PennyLane-specific operations while removing unrelated helper code, invalid transformations, and near-duplicate entries.

\textbf{Stage 1: Extraction.}
We first retrieve complete Python source files using direct GitHub URL access. Preliminary trials with pre-chunked CSV-based sources led to severe fragmentation and retained only 0.6\% of usable code, mainly because function boundaries, imports, and surrounding execution context were often lost. We therefore operate directly on full source files and perform structural parsing.

Function-level extraction is performed using Python AST analysis. Each file is parsed into an abstract syntax tree, and candidate functions are retained when they contain PennyLane-specific operations or quantum-programming structure. This allows the pipeline to preserve syntactic scope, argument lists, return statements, and internal control flow while filtering out unrelated helper functions. After this stage, 11,643 functions are retained, corresponding to a 76.9\% extraction retention rate.

\textbf{Stage 2: Verification.}
Extracted functions are passed through a four-layer verification pipeline to ensure that the resulting code remains syntactically valid, structurally coherent, and consistent with the original quantum operations. The verification layers are:
\begin{enumerate}
  \item \textbf{Syntax validation:} each candidate is checked using Python compilation.
  \item \textbf{Import validation:} required libraries and quantum-programming dependencies must remain present.
  \item \textbf{Quantum-operation preservation:} gate count changes must remain within 20\%, measurements preserved, and \texttt{qml.*} call changes limited to at most 50\%.
  \item \textbf{Semantic structure preservation:} the function must preserve its return structure and intended output path.
\end{enumerate}
We adopt a conservative fallback policy: if any verification step fails, the original extracted code is retained instead of the transformed version. After verification, 11,344 functions remain, corresponding to a 97.4\% verification retention rate.

\textbf{Stage 3: Deduplication.}
We apply MinHash LSH~\cite{leskovec2020minhash} with 128 permutations and a 70\% similarity threshold to remove near-duplicate samples from the verified GitHub-derived corpus. This stage produces 10,179 deduplicated GitHub-derived instruction-code pairs. We then append 3,210 QHack and tutorial archive entries, resulting in the final \textbf{PennySynth-13K} corpus of 13,389 verified instruction-code pairs.

\subsubsection{Prompt Generation}

Each code sample in PennySynth-13K is paired with a natural-language instruction to support natural-language-to-code retrieval. Since many extracted functions do not contain an explicit task description, we generate instructions automatically using the Gemini API under a constrained template. The template requires a short imperative description of 20-40 words and asks the model to mention the main PennyLane operations, device or wire information when available, and the expected return type. 

To assess whether the generated instructions are suitable for retrieval, we perform an internal retrieval consistency check. Each generated instruction is used as a query against the constructed corpus, and the retrieval is considered correct when the original code sample appears among the retrieved candidates. Under this evaluation, the generated instructions achieve 95\% Top-1 retrieval accuracy and 100\% Top-5 retrieval accuracy, indicating that the instructions preserve enough task-specific information to recover their corresponding code samples.

\subsection{Embedding Strategy}

The retrieval stage requires an embedding model that can align natural-language task descriptions with PennyLane code examples. We evaluate two configurations in the ablation study. The \textbf{MiniLM configuration} uses \texttt{all-MiniLM-L6-v2}, a 384-dimensional general-purpose sentence embedding model, serving as a lightweight semantic retrieval baseline. The \textbf{PennySynth configuration} uses \texttt{st-codesearch-distilroberta-base}, a 768-dimensional model trained on CodeSearchNet for natural-language-to-code retrieval, making it better suited for matching QHack problem descriptions to instruction-code pairs.

Given an expanded query $q'$ and a candidate document $d_i$, retrieval similarity is computed using cosine similarity between their embeddings:
\begin{equation}
  s(q', d_i) = \frac{\mathbf{e}_{q'} \cdot \mathbf{e}_i}{\|\mathbf{e}_{q'}\| \cdot \|\mathbf{e}_i\|},
  \label{eq:cosine}
\end{equation}
where $\mathbf{e}_{q'}$ represents the embedding of the expanded challenge query, and $\mathbf{e}_i$ is the embedding of the $i$-th instruction-code pair in the knowledge base.

In our retrieval consistency evaluation, MiniLM yields an average cosine similarity of approximately $\bar{s}\approx0.45$, while the PennySynth configuration increases this value to $\bar{s}=0.726$. We therefore use the PennySynth configuration as the default retrieval backbone in the main experiments.

\subsection{Retrieval and Prompt Design}
Given a challenge description $q$, the system first performs query expansion to produce a more focused PennyLane-oriented query $q'$. The expanded query is embedded using the selected retrieval backbone and compared against the embedded instruction-code pairs in the ChromaDB knowledge base using Eq.~\eqref{eq:cosine}.

The system retrieves the top-$k=5$ most similar examples. If the maximum similarity score among the retrieved candidates falls below the threshold $\tau=0.60$, the system falls back to base generation to prevent \emph{context contamination}, i.e., the inclusion of irrelevant or weakly related examples that may mislead the model. When the threshold is met, the prompt is augmented with the retrieved examples and an explicit selective-context instruction: \emph{``If retrieved examples are not relevant to this challenge, ignore them and rely on your own PennyLane knowledge.''}

After generation, the produced code is executed against the challenge test cases. If execution fails, the model enters an auto-fix loop that provides the previous code, execution feedback, and the original challenge to a repair prompt, with at most $T=2$ repair attempts. Algorithm~\ref{alg:rag} summarizes this procedure.

\begin{figure}[t!]
\begin{algorithm}[H]
\caption{PennySynth Inference Pipeline}
\label{alg:rag}
\begin{algorithmic}[1]
\Require Challenge $q$, knowledge base $\mathcal{D}$,
         threshold $\tau$, max fixes $T$
\Ensure Solution $c^*$, result $r^*$
\State $q' \gets \text{LLM}_{\mathrm{expand}}(q)$
\State $\mathcal{R} \gets \text{Top-}k\text{-retrieve}(
       \text{Enc}(q'), \mathcal{D})$
\State $p \gets \text{BasePrompt}(q)$ \textbf{ if}
       $\max_{d \in \mathcal{R}} s < \tau$
       \textbf{ else} $\text{RAGPrompt}(q, \mathcal{R})$
\State $c_0 \gets \text{ExtractCode}(\text{LLM}(p))$;
       $r_0 \gets \text{Execute}(c_0)$
\For{$t = 1,\ldots,T$}
  \If{$r_{t-1}.\text{pass}$} \textbf{break} \EndIf
  \State $c_t \gets \text{ExtractCode}(
         \text{LLM}_{\mathrm{fix}}(c_{t-1}, r_{t-1}, q, \mathcal{R}))$
  \State $r_t \gets \text{Execute}(c_t)$
\EndFor
\State \Return $c_t, r_t$
\end{algorithmic}
\end{algorithm}
\end{figure}

\subsection{Quantum-Adapted CodeBLEU}
\label{sec:codebleu}
We build on CodeBLEU to evaluate generated PennyLane code while accounting for quantum-specific API usage. Standard CodeBLEU combines token-level similarity, weighted n-gram matching, AST matching, and dataflow matching, but it does not explicitly distinguish PennyLane operations from generic Python tokens. This distinction is important because \texttt{qml.*} operations, device declarations, and measurement statements directly affect circuit behavior and execution correctness.

We introduce two quantum-specific modifications. First, \textbf{token upweighting}: all \texttt{qml.*} tokens are repeated $3\times$ before computing the weighted n-gram component, giving greater emphasis to PennyLane-specific operations. Second, we replace the standard dataflow component with a \textbf{quantum dataflow match} based on Jaccard similarity over the set $K(c)$ extracted from a code sample $c$, where $K(c)$ contains \texttt{qml.*} gate names, device types, and measurement return types:
\begin{equation}
\text{DF}(h,r) = \frac{|K(h) \cap K(r)|}{|K(h) \cup K(r)|},
\label{eq:df}
\end{equation}
where $h$ is the code generated by the LLM, also referred to as the hypothesis, and $r$ is the QHack challenge template used as the reference. Thus, $\text{DF}(h,r)$ measures the overlap between the quantum-relevant operations, device types, and measurement structures extracted from the generated code and the reference template.
%% ─────────────────────────────────────────────────────────────
\section{Experiment Settings}

\subsection{Benchmark}

We evaluate PennySynth on QHack, a PennyLane coding competition covering three benchmark years: 25 challenges from 2022, 28 from 2023, and 21 from 2024, for a total of 74 challenges. Following QHackBench~\cite{basit2025qhackbench}, we use these tasks as a standardized benchmark for executable PennyLane code generation. Each challenge provides a problem statement, a code template to complete, input-output specifications, and test functions for validating submitted solutions. Generated code is considered correct only when it passes the corresponding tests under this execution-based evaluation protocol.
\subsection{Systems}

We evaluate six recent LLMs as base generators without retrieval to establish strong baselines: \textbf{Claude Sonnet 4.6}~\cite{anthropic2024claude}, \textbf{GPT-5.5}~\cite{openai2024gpt45}, \textbf{Gemini 2.5 Pro}~\cite{google2024gemini}, \textbf{Qwen3-235B-A22B}~\cite{qwen3_2025}, \textbf{GLM-5.1}~\cite{glm2024}, and \textbf{DeepSeek-V3}~\cite{deepseek2024}. 

% PennySynth augments the two strongest base models with the full RAG pipeline using PennySynth-13K and code-aware embeddings.

\subsection{Metrics}

We report four evaluation metrics. \textbf{pass@5} measures whether at least one of five independently generated solutions passes all test cases for a given challenge. We also report \textbf{pass@1} for single-generation evaluation of the RAG systems. \textbf{Partial credit} measures the fraction of test cases passed within each challenge. \textbf{CodeBLEU} and \textbf{ROUGE-L} measure code similarity against the available QHack 2022 reference templates, with CodeBLEU evaluated using the quantum-adapted variant described in Section~\ref{sec:codebleu}. \textbf{Hallucination rate} measures the percentage of generated solutions containing \texttt{qml.*} calls absent from a curated PennyLane operation whitelist.

\subsection{Implementation}

All generation experiments use a temperature of 0.7 and a maximum generation length of 3,000 tokens. Retrieval is performed with top-$k{=}5$, while a relevance threshold of $\tau{=}0.60$ is used to exclude weakly related examples that may introduce noise into the prompt. The auto-fix loop permits up to $T{=}2$ retry attempts.
%% ─────────────────────────────────────────────────────────────

\section{Results and Discussion}

\subsection{Multi-Model Baseline Comparison}

\begin{figure}[ht!]
\centering
\includegraphics[width=\columnwidth]{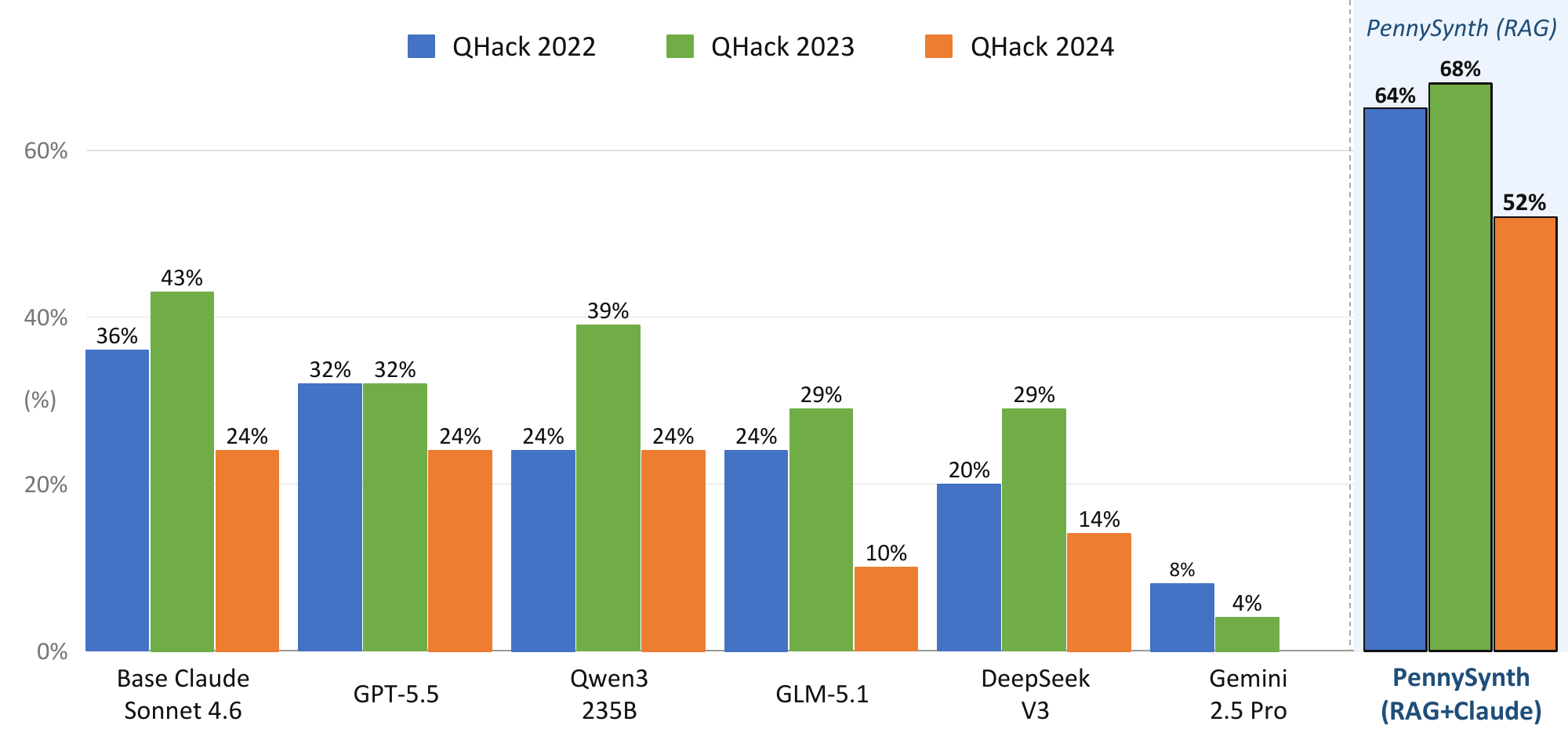}
\caption{pass@5 across six base LLMs and PennySynth (RAG+Claude) on QHack 2022, 2023, and 2024. PennySynth substantially outperforms all base LLMs.} %despite being evaluated under the stricter single-generation metric.}
\label{fig:pass5}
\end{figure}

We first evaluate PennySynth on QHack challenges from 2022, 2023, and 2024 and compare it against six recent base LLMs without retrieval. This evaluation establishes the main performance gap between general-purpose code generation and retrieval-augmented PennyLane code generation. As shown in Fig.~\ref{fig:pass5}, PennySynth achieves the strongest performance across all three benchmark years, reaching 64.0\%, 68.0\%, and 52.0\% pass@5 on QHack 2022, 2023, and 2024, respectively.

Among the base models, Claude Sonnet 4.6 obtains the best or tied-best results across the three years, with pass@5 scores of 36.0\%, 42.9\%, and 23.8\%. GPT-5.5 is competitive on QHack 2022, while Qwen3-235B-A22B approaches Claude on QHack 2023. However, all base models remain far below PennySynth. Relative to Claude Sonnet 4.6 without retrieval, PennySynth improves pass@5 by +28.0, +25.1, and +28.2 percentage points on QHack 2022, 2023, and 2024, respectively. These gains indicate that retrieval from PennySynth-13K provides task-specific PennyLane patterns that are not consistently recovered from the model's internal knowledge alone.

%%%comparaison 
To position PennySynth against the closest prior PennyLane-oriented code-generation baseline, we compare it with QHackBench, which also reports retrieval-augmented results on QHack challenges. On the overlapping QHack years, PennySynth achieves 68.0\% pass@5 on QHack 2023 and 52.0\% on QHack 2024, compared with 60.7\% and 33.3\% reported by QHackBench RAG~\cite{basit2025qhackbench}. This corresponds to gains of +7.3 and +18.7 percentage points, respectively. These improvements indicate that the proposed PennySynth-13K dataset and retrieval pipeline provide stronger support for competition-style PennyLane code generation than the prior QHackBench RAG setting.

\subsection{PennySynth RAG Performance}

We next evaluate PennySynth-13K as a retrieval component across different generator models. Table~\ref{tab:pennysynth} reports pass@1 and pass@5 results for seven LLMs on QHack 2022-2024 using the same retrieval pipeline. This analysis examines whether retrieved PennyLane examples improve generation across model families and motivates the choice of Claude Sonnet 4.6 as the main generator for the remaining experiments.

\begin{table}[htbp]
  \centering
  \caption{Performance of RAG with PennySynth-13k across 7 cutting-edge models (pass@1 / pass@5) on QHack challenges from 2022 to 2024.}
  \label{tab:pennysynth}
  \renewcommand{\arraystretch}{1.3}
  \begin{tabular}{l *{6}{c}}
    \toprule
    \multirow{2}{*}{\textbf{Model}}
      & \multicolumn{2}{c}{\textbf{2022}}
      & \multicolumn{2}{c}{\textbf{2023}}
      & \multicolumn{2}{c}{\textbf{2024}} \\
    \cmidrule(lr){2-3} \cmidrule(lr){4-5} \cmidrule(lr){6-7}
      & 1 & 5 & 1 & 5 & 1 & 5 \\
    \midrule
    Claude Sonnet 4.6
      & 56\% & 64\% & \cellcolor{bestcell}\textbf{61\%} & \cellcolor{bestcell}\textbf{68\%}
      & \cellcolor{bestcell}\textbf{43\%} & 52\% \\
    GPT-5.5
      & \cellcolor{bestcell}\textbf{64\%} & \cellcolor{bestcell}\textbf{68\%} & 50\% & 57\%
      & 29\% & 43\% \\
    Qwen3-235B
      & 52\% & 64\% & 50\% & 61\%
      & \cellcolor{bestcell}\textbf{43\%} & \cellcolor{bestcell}\textbf{53\%} \\
    GLM-5.1
      & 20\% & 48\% & 11\% & 32\%
      & 5\%  & 14\% \\
    DeepSeek-V3
      & 32\% & 48\% & 36\% & 36\%
      & 19\% & 33\% \\
    Gemini 2.5 Pro
      & 8\%  & 8\%  & 4\%  & 4\%
      & 0\%  & 0\%  \\
    LLaMA 3.1-8B
      & 12\% & 20\% & 11\% & 32\%
      & 0\%  & 0\%  \\
    \bottomrule
  \end{tabular}
\end{table}

The results indicate that PennySynth-13K is most effective when paired with strong generator models. Claude Sonnet 4.6 provides the most stable performance across the three benchmark years, achieving 64\%, 68\%, and 52\% pass@5 on QHack 2022, 2023, and 2024, respectively. GPT-5.5 obtains the best result on QHack 2022, while Qwen3-235B achieves the highest pass@5 on QHack 2024. These findings show that the retrieval pipeline is not tied to a single model, although its effectiveness depends on the generator's ability to adapt retrieved examples to the target challenge. 

The results also reveal a clear difference between strong and weaker generators. GLM-5.1 and DeepSeek-V3 obtain moderate but less stable performance, while Gemini 2.5 Pro and LLaMA 3.1-8B remain weak across most settings. This suggests that retrieval alone is not sufficient for reliable PennyLane code generation. The model must still follow strict output formats, use the PennyLane API correctly, and modify retrieved examples instead of copying incompatible code. Based on its stable performance, Claude Sonnet 4.6 is used as the main PennySynth generator in the following analyses.

\subsection{Code Quality Analysis}

Beyond functional correctness, we evaluate whether generated solutions preserve the expected structure of the official QHack reference code. Although official reference code is available for all benchmark years, we focus this code-level analysis on QHack 2022 to provide a controlled comparison over a fixed set of challenges. In this setting, the reference templates define the required structure, while the model must complete the missing functions correctly. Table~\ref{tab:codebleu1} reports CodeBLEU, ROUGE-L, AST match, and quantum dataflow match on the selected QHack 2022 challenges.

\begin{table}[htbp]
\centering
\caption{Code quality against official QHack reference templates on 11 selected QHack 2022 challenges. 
PennySynth and RAG-MiniLM use Claude Sonnet 4.6 as the generator, while Base Gemini uses Gemini 2.0 Flash. CB = CodeBLEU, RL = ROUGE-L, AST = AST match, and DF = quantum dataflow match.}
\label{tab:codebleu1}
\setlength{\tabcolsep}{4pt}
\begin{tabular}{lcccc}
\toprule
\textbf{System (Generator Model)} & \textbf{CB$\uparrow$} & \textbf{RL$\uparrow$} & \textbf{AST$\uparrow$} & \textbf{DF$\uparrow$} \\
\midrule
Gemini 2.0 Flash (no RAG)       & 0.531 & 0.569 & 0.685 & 0.548 \\
RAG-MiniLM (Claude Sonnet 4.6)       & 0.495 & 0.473 & 0.688 & 0.573 \\
Claude Sonnet 4.6 (no RAG)      & 0.480 & 0.448 & 0.696 & 0.575 \\
\textbf{PennySynth (Claude Sonnet 4.6)} & \textbf{0.522} & \textbf{0.529} & 0.688 & 0.570 \\
\bottomrule
\end{tabular}
\end{table}

The results show that similarity to the official reference template does not always align with functional correctness. Gemini 2.0 Flash obtains the highest CodeBLEU score, but its functional pass rate remains low, indicating that a generated solution can preserve much of the expected template structure while still completing the missing functions incorrectly or producing incorrect numerical outputs. In contrast, PennySynth improves over Claude Sonnet 4.6 without retrieval in both CodeBLEU and ROUGE-L, increasing CodeBLEU from 0.480 to 0.522 and ROUGE-L from 0.448 to 0.529. These gains suggest that retrieval helps the generator use more appropriate PennyLane tokens, function patterns, and API-level expressions within the required template.

However, the AST and quantum dataflow scores remain close across Claude-based systems. This indicates that retrieval mainly improves local code formulation and API alignment rather than substantially changing the high-level circuit structure constrained by the template. Therefore, code-similarity metrics should be interpreted as supporting evidence rather than substitutes for functional evaluation. For this reason, pass@k remains the primary metric in our benchmark.

\subsection{Ablation Studies}
We conduct ablation experiments on QHack 2022 to identify which components of PennySynth-13K contribute most to retrieval performance. The analysis first separates the effect of the embedding model from the effect of dataset size, then examines the contribution of each source category in the full corpus.

\subsubsection{Embedding Model and Dataset Composition}
\label{sec:ablation}

To disentangle the effects of the code-aware embedding model and the expanded dataset, we conduct a controlled 2$\times$2 ablation on QHack 2022 while keeping all other hyperparameters fixed. Table~\ref{tab:ablation} compares four configurations across two embedding models and two corpus sizes.

\begin{table}[htpbt]
\centering
\caption{Ablation study on QHack 2022 (25 challenges). MiniLM = \texttt{all-MiniLM-L6-v2} (384-dim); PennySynth emb.\ = \texttt{st-codesearch-distilroberta-base} (768-dim). Dataset sizes: Community = 10,179 pairs (unofficial + official GitHub); Full = 13,389 pairs (+ QHack archives). See Table~\ref{tab:profiling} for source breakdown.}
\label{tab:ablation}
\setlength{\tabcolsep}{3pt}
\begin{tabular}{llcc}
\toprule
\textbf{Embedding} & \textbf{Dataset} & \textbf{pass@1} & \textbf{Data effect} \\
\midrule
MiniLM (384-dim)        & Community (10K) & 48\% & \multirow{2}{*}{+0 pp} \\
MiniLM (384-dim)        & Full (13K)      & 48\% & \\
\midrule
PennySynth emb. (768-dim) & Community (10K) & 52\% & \multirow{2}{*}{+4 pp} \\
\textbf{PennySynth emb. (768-dim)} & \textbf{Full (13K)} & \textbf{56\%} & \\
\midrule
Embedding effect (10K)  & ---             & +4 pp & \\
Embedding effect (13K)  & ---             & +8 pp & \\
\bottomrule
\end{tabular}
\end{table}

Three findings emerge. First, the code-aware embedding is the primary performance driver: switching from MiniLM to the PennySynth embedding model delivers +4~pp at the smaller dataset size and +8~pp at the full dataset, confirming that cross-modal code retrieval quality gates all downstream gains. Second, dataset scale is a conditional contributor: expanding from 10K to 13K improves pass@1 only under the code-aware embedding (+4~pp), while providing no benefit under the general-purpose embedding. This occurs because MiniLM cannot distinguish between the additional competition-style entries and existing similar entries, introducing retrieval noise that cancels coverage gains. Third, the two factors are synergistic: the combined effect (+8~pp) exceeds the sum of individual contributions, as the code-aware embedding is better positioned to exploit the structural diversity introduced by the larger corpus.

Challenge-level analysis further reveals that although MiniLM achieves identical pass@1 at both dataset sizes, the two configurations solve different challenges: 7 are solved by both, 3 only by the smaller dataset (algorithm challenges favoring focused retrieval), and 4 only by the larger dataset (chemistry and QML challenges benefiting from broader coverage). The union pass rate of 14/25 (56\%) matches the PennySynth result, confirming that dataset expansion under MiniLM trades precision for coverage without net gain.

\subsubsection{Dataset Profiling}
\label{sec:profiling_ablation}

To assess the contribution of each source category in PennySynth-13K, we evaluate individual and cumulative corpus configurations using the PennySynth embedding model. Table~\ref{tab:profiling} reports the contribution of unofficial GitHub code, official PennyLane repositories, and QHack/tutorial archives.

\begin{table}[htpbt]
\centering
\caption{Dataset profiling ablation on QHack 2022 (25 challenges)
  using PennySynth embedding. Individual source rows show each corpus
  in isolation; cumulative rows show incremental gains.
  Gain is relative to the previous cumulative row.}
\label{tab:profiling}
\setlength{\tabcolsep}{3pt}
\begin{tabular}{lrcc}
\toprule
\textbf{Corpus Configuration} & \textbf{Size} & \textbf{pass@1} & \textbf{Gain} \\
\midrule
\multicolumn{4}{l}{\textit{Individual sources (isolated)}} \\
Unofficial GitHub only              &  8,245 & 40\% & --- \\
Official PennyLane only             &  1,934 & 48\% & --- \\
QHack archives only     &  3,210 & 48\% & --- \\
\midrule
\multicolumn{4}{l}{\textit{Cumulative (incremental addition)}} \\
Unofficial GitHub only              &  8,245 & 40\% & ---    \\
+ Official PennyLane repos          & 10,179 & 52\% & +12 pp \\
+ QHack archives        & 13,389 & 56\% & +4 pp  \\
\bottomrule
\end{tabular}
\end{table}

This presents both individual and cumulative source contributions. When evaluated in isolation, unofficial GitHub (40\%), official PennyLane repos (48\%), and QHack archives (48\%) each provide moderate standalone performance, confirming that no single source alone is sufficient. The cumulative analysis reveals a clear incremental structure: unofficial GitHub alone achieves 40\%, adding official PennyLane repositories (+1,934 entries) produces the largest single gain of +12~pp (40\%~$\to$~52\%), and adding QHack and tutorial archives contributes a further +4~pp (52\%~$\to$~56\%). 

The disproportionate +12~pp gain from official repositories, despite their smaller size (1,934 vs 8,245 unofficial entries), confirms that source quality and canonical API correctness matter more than raw corpus scale. Official repositories provide version-consistent, idiomatic PennyLane implementations that directly address the hallucination and API misuse failure modes identified in Table~\ref{tab:errors}. The synergistic effect, where the full 13K corpus outperforms any individual source, validates the deliberate multi-source design of PennySynth-13K.

\begin{table}[ht!]
\centering
\caption{Error profile on QHack 2024 as percentage of each system's total failures. Base Claude: $n=11$ failures, PennySynth: $n=12$ failures.}
\label{tab:errors}
\setlength{\tabcolsep}{4pt}
\begin{tabular}{lccc}
\toprule
\textbf{Category} & \textbf{Base} & \textbf{PennySynth} & \textbf{$\Delta$} \\
\midrule
\texttt{formatting\_failure} & 45\% & 42\% & $-$3\% \\
\texttt{hallucination}       & 27\% & 33\% & +6\% \\
\texttt{reasoning\_error}    & 18\% &  8\% & $-$10\% \\
\texttt{api\_misuse}         &  9\% & 17\% & +8\% \\
\texttt{timeout}             &  0\% &  0\% & 0\% \\
\bottomrule
\end{tabular}
\end{table}

\subsection{Failure Analysis}

After the QHack 2022 code-quality and ablation analyses, we examine the remaining failures on QHack 2024 to better understand where retrieval still falls short. Table~\ref{tab:errors} reports the error distribution for Claude Sonnet 4.6 without retrieval and PennySynth. Errors are grouped into formatting failures, hallucinations, reasoning errors, API misuse, and timeouts.

Formatting failures dominate both profiles at 45\% and 42\%, caused by non-compliance with QHack's strict template structure. The key shift introduced by retrieval augmentation is a trade-off between reasoning errors and context-induced errors: PennySynth reduces reasoning errors from 18\% to 8\% while hallucination increases from 27\% to 33\% and API misuse rises from 9\% to 17\%. Retrieved context occasionally surfaces gate names from older PennyLane versions or incompatible function signatures, introducing the same mechanism that supplies correct domain-specific patterns also surfaces incorrect ones at moderate similarity.

This indicates that retrieved context can introduce outdated gate names, incompatible PennyLane syntax, or function signatures that do not match the target challenge. Thus, the same mechanism that provides useful domain-specific patterns can also introduce errors when the retrieved example is only partially aligned with the task.

No timeout failures are observed for either system, suggesting that the main limitations are not execution cost but formatting, API compatibility, and challenge-specific reasoning. 

\subsection{Dataset Pipeline Case 1: Ground Transformation}
\label{sec:casestudy_data}

We trace Entry~\#32 from PennySynth-13K through every stage of the pipeline, showing the actual prompts, LLM responses, and transformation decisions at each step. All outputs were captured by running the pipeline scripts on the actual dataset entry.

\subsubsection*{Stage 0: Raw GitHub File}
 
The GitHub Search API returns a CSV of PennyLane-related Python files. The pipeline fetches raw file content via GitHub URL. For Entry~\#32, the source is \texttt{quantum\_classifier.py}, a 36-line community repository file containing 7~functions:

\begin{itemize}
  \item[\textcolor{rejectred}{\ding{55}}] 5 non-quantum helpers: \texttt{normalize\_data}, \texttt{plot\_loss}, \texttt{accuracy}, \texttt{train\_epoch}, \texttt{forward}
  \item[\textcolor{retaingreen}{\ding{51}}] 1 quantum function: \texttt{quantum\_circuit}
\end{itemize}

\subsubsection*{Stage 1: AST Extraction \normalfont\texttt{(stage1.py)}}
 
The AST parser scans all 7~functions and classifies each using the \texttt{lenient} retention strategy. Functions containing any \texttt{qml.*} call are classified as \texttt{direct}; the remaining 5~helpers are rejected. The extracted function:
 
\begin{lstlisting}[
  style=pythoncode,
  caption={Entry \#32 Extracted quantum function (pre-modernization)},
  label={lst:entry32_raw}
]
def quantum_circuit(inputs, weights):
    qml.templates.AngleEmbedding(inputs,
        wires=range(n_qubits))
    qml.templates.
        StronglyEntanglingLayers(weights,
        wires=range(n_qubits))
    return [qml.expval(qml.PauliZ(i))
            for i in range(n_qubits)]
\end{lstlisting}
 
\noindent\textbf{Extraction result:} 2 of 7 functions retained (\textcolor{rejectred}{71\% rejection} of non-quantum code).

\subsubsection*{Stage 2: LLM Transformation \normalfont\texttt{(Stage2\_Verified.py)}}
 
The extracted code is sent to \texttt{claude-3.5-haiku} via OpenRouter. The system prompt instructs the model to fix deprecated APIs while preserving all quantum operations:
 
\begin{systemprompt}
\small\itshape
You are a PennyLane code quality assistant. Fix deprecated APIs only. Return clean Python code with no markdown fences.
\end{systemprompt}
 
\begin{userprompt}
\small\itshape
Fix this PennyLane code by replacing deprecated \texttt{qml.templates.*} calls with their modern equivalents.\\[4pt]
\textbf{Rules:}
\begin{enumerate}[nosep, leftmargin=1.2em]
  \item Replace \texttt{qml.templates.AngleEmbedding} $\rightarrow$ \texttt{qml.AngleEmbedding}
  \item Replace \texttt{qml.templates.Strongly-\\EntanglingLayers} $\rightarrow$ \texttt{qml.StronglyEntanglingLayers}
  \item Keep ALL other code identical
  \item Add missing imports at the top
  \item Return ONLY the fixed Python code
\end{enumerate}
\textbf{Code:} \texttt{[function body]}
\end{userprompt}
 
\medskip
\noindent Claude Haiku responds with the modernized code:
 
\begin{lstlisting}[
  style=pythoncode,
  caption={Entry \#32 Modernized output from Claude Haiku},
  label={lst:entry32_modern}
]
import pennylane as qml

def quantum_circuit(inputs, weights):
    qml.AngleEmbedding(inputs, wires=range(n_qubits))
    qml.StronglyEntanglingLayers(weights,
        wires=range(n_qubits))
    return [qml.expval(qml.PauliZ(i))
            for i in range(n_qubits)]
\end{lstlisting}
 
\noindent\textbf{Operation preservation check:}

\texttt{qml.templates.*}~$2{\to}0$~\textcolor{retaingreen}{\ding{51}}\enspace
Modern API~$0{\to}2$~\textcolor{retaingreen}{\ding{51}}\enspace
\texttt{qml.*} total~$4{\to}4$~\textcolor{retaingreen}{\ding{51}}\enspace
Measurements~$1{\to}1$~\textcolor{retaingreen}{\ding{51}}

\smallskip
\noindent\textbf{Verdict:} \colorbox{retaingreen!12}{\texttt{\textcolor{retaingreen}{transformed\_valid}}}

\subsubsection*{Stage 3: MinHash Deduplication \normalfont\texttt{(Stage3\_Dedup.py)}}
 
The modernized code is tokenized into 3-shingles using \texttt{datasketch} MinHash (128~permutations). Entry~\#265\, a structurally similar classifier from a different repository is compared:
 
\smallskip
\noindent\textbf{Deduplication check (Entry \#32 vs.\ \#265):}
Shingles: $48$ vs.\ $89$\enspace$\mid$\enspace
Intersection/Union: $43/94$\enspace$\mid$\enspace
Jaccard: $0.457 < 0.70$\enspace$\Rightarrow$\enspace
\colorbox{retaingreen!12}{\textcolor{retaingreen}{\textbf{UNIQUE $\Rightarrow$ RETAIN}}}
 
\smallskip
\noindent Both entries are retained since their implementation differences (device declaration style, variable scoping) produce sufficient token divergence despite semantic similarity.

\subsubsection*{Stage 4: Instruction Generation \normalfont\texttt{(PromptGen\_Hybrid.py)}}
 
The final code is sent to \texttt{gemini-2.5-flash-lite} with the instruction-style prompt:
 
\begin{userprompt}
\small\itshape
Based on this PennyLane code, write a clear instruction prompt (20--40 words).\\[2pt]
\textbf{Format:} Start with an action verb. Include: key quantum operations, number of qubits/wires, return type.\\[2pt]
Return ONLY the instruction prompt.
\end{userprompt}
 
\noindent\textbf{Generated instruction:}

\begin{assistantprompt}
\small\itshape
Implement a PennyLane quantum circuit with \texttt{n\_qubits} using \texttt{AngleEmbedding} and \texttt{StronglyEntanglingLayers}, returning expectation values of PauliZ on each wire.
\end{assistantprompt}
 
\subsubsection*{Quality Improvement Summary}
 
The pipeline transformed a raw community file containing 5~irrelevant helper functions and 2~deprecated API calls into a single verified, modernized, instruction-annotated entry. The key transformation\,---\,\texttt{qml.templates.AngleEmbedding} $\rightarrow$ \texttt{qml.AngleEmbedding}\, directly reduces the API misuse failure mode identified in Table~\ref{tab:errors}, since retrieved code using the old namespace causes \texttt{AttributeError} at inference time under PennyLane~$\geq$v0.40.

\subsection{Dataset Pipeline Case 2: Precise Deduplication}
\label{sec:casestudy_data2}

This case demonstrates that the deduplication stage is \emph{precision-preserving}: structurally similar entries implementing meaningfully different circuits are correctly retained as distinct knowledge base entries.

\subsubsection*{Entry \#265 Candidate Near-Duplicate of Entry \#32}
 
During Stage~3 deduplication, Entry~\#265 is compared against Entry~\#32:
 
\begin{lstlisting}[
  style=pythoncode,
  caption={Entry \#265 Candidate near-duplicate with dynamic device construction},
  label={lst:entry265}
]
@qml.qnode(create_quantum_device(n_qubits),
           interface='torch')
def qnode(inputs, weights):
    qml.AngleEmbedding(inputs,
        wires=range(n_qubits))
    qml.StronglyEntanglingLayers(weights,
        wires=range(n_qubits))
    return [qml.expval(qml.PauliZ(i))
            for i in range(n_qubits)]
\end{lstlisting}
 
\subsubsection*{Pipeline Stages for Entry \#265}
 
\textbf{Stage 1 (Extraction):}
The AST parser extracts this function from a different community repository. Classified as \texttt{direct} (3~\texttt{qml.*} calls). No deprecated API is present, Entry~\#265 already uses the modern \texttt{qml.AngleEmbedding} namespace.
 
\medskip
\textbf{Stage 2 (Verification):}
All four verification layers pass. No transformation required. Verdict: \colorbox{retaingreen!12}{\texttt{\textcolor{retaingreen}{original\_valid}}}
 
\medskip
\textbf{Stage 3 Deduplication Decision:}
 
MinHash LSH (128~permutations, $k{=}3$ shingles, threshold${=}0.70$) computes the following:
 
\noindent\textbf{Deduplication check (Entry \#32 vs.\ \#265):}
Shingles: $48$ vs.\ $89$\enspace$\mid$\enspace
Intersection/Union: $43/94$\enspace$\mid$\enspace
Jaccard: $0.457 < 0.70$\enspace$\Rightarrow$\enspace
\colorbox{retaingreen!12}{\textcolor{retaingreen}{\textbf{UNIQUE $\Rightarrow$ RETAIN}}}
 
\smallskip
\noindent Despite both entries implementing \texttt{AngleEmbedding} + \texttt{StronglyEntanglingLayers} with PauliZ measurements, they differ structurally:
 
\noindent Despite both entries implementing the same gate set, they differ structurally:
Entry~\#32 declares the device globally (\texttt{dev = qml.device(...)}), uses 4~fixed wires, and names the function \texttt{quantum\_circuit};
Entry~\#265 wraps device creation in a helper (\texttt{create\_quantum\_device(n\_qubits)}), uses a variable wire count, names the function \texttt{qnode}, and specifies \texttt{interface='torch'} in the decorator.
These differences produce sufficient token divergence to fall below the 0.70 threshold.
 
\subsubsection*{Stage 4 Instruction Generation}
 
\begin{userprompt}
\small\itshape
Based on this PennyLane code, write a clear instruction prompt (20--40 words).\\[2pt]
\textbf{Format:} Start with an action verb. Include: key quantum operations, number of qubits/wires, return type.\\[2pt]
Return ONLY the instruction prompt.\\[2pt]
\textbf{Code:} \texttt{[function body]}
\end{userprompt}
 
\noindent\textbf{Generated instruction:}

\begin{assistantprompt}
\small\itshape
Implement a PennyLane QNode using a dynamically created \texttt{default.qubit} device with \texttt{n\_qubits} wires, applying \texttt{AngleEmbedding} and \texttt{StronglyEntanglingLayers}, and returning PauliZ expectation values.
\end{assistantprompt}
 
\subsubsection*{Value of Retaining Both Entries}
 
At retrieval time, Entry~\#32 and Entry~\#265 serve different challenge patterns:
\begin{itemize}[leftmargin=1.5em]
  \item \textbf{Entry \#32} matches challenges with \emph{fixed-wire device declarations} (common in QHack 2022 templates).
  \item \textbf{Entry \#265} matches challenges requiring \emph{dynamic device construction} with variable qubit counts.
\end{itemize}
\noindent Retaining both increases retrieval coverage across circuit topology variants, precisely the complementary coverage that the profiling ablation in Section~\ref{sec:profiling_ablation} confirms is valuable. Over-aggressive deduplication (e.g., threshold${}=0.40$) would have incorrectly removed Entry~\#265, reducing coverage for dynamic-device challenges.

\subsection{Retrieval Case Studies}

\paragraph{Successful retrieval: \texttt{three\_shipping\_companies}}
This challenge requires a neutral-atom gate set using $\text{RY}(\pi/2)$ and CZ gates, absent from standard PennyLane tutorials. PennySynth retrieved a near-identical QHack 2023 challenge at similarity 0.850, enabling the model to adapt the correct gate set. Cross-system CodeBLEU of 0.237 confirms fundamentally different code was produced compared to the base model, with only the retrieval-augmented version succeeding.

\paragraph{Retrieval interference: \texttt{GHZ\_inn}}
Base Claude solves this correctly while PennySynth fails despite a cross-system CodeBLEU of 0.752. Retrieved context introduced a subtly different circuit topology that overrode the model's correct internal reasoning, a failure mode when retrieved similarity is above threshold but the example targets a different problem variant.

\paragraph{Coverage gap: \texttt{mojito\_hhlime\_twist}}
Both systems fail with cross-system CodeBLEU of 0.016, indicating entirely different solution strategies. Neither parametric knowledge nor retrieved examples contained sufficient information, representing a genuine capability gap motivating Graph RAG as future work.

%% ─────────────────────────────────────────────────────────────
\section{Conclusion}

We presented PennySynth, a retrieval-augmented generation system for automated PennyLane quantum code synthesis built around an automated three-stage data synthesis pipeline producing PennySynth-13K, a corpus of 13,389 AST-verified instruction-code pairs. By adopting a code-aware embedding model and expanding the knowledge base with competition-style QHack archives, PennySynth achieves up to 68\% pass@5 on QHack challenges, a +25~pp improvement over a strong base LLM. A controlled 2$\times$2 ablation demonstrates that code-aware embeddings are the primary performance driver, with dataset expansion providing synergistic gains only under precise retrieval. Multi-model evaluation across six state-of-the-art LLMs establishes the first comprehensive baseline for PennyLane competition-level code generation.

Future work will address: (1)~a custom PennyLane knowledge graph for Graph RAG; (2)~the combined PennySynth~+~fine-tuned system; (3)~hybrid dense-sparse retrieval (BM25 + CodeBERT); and (4)~extending the multi-model evaluation to additional LLMs and newer QHack benchmark years.
\section*{Data Availability}
The proposed dataset is publicly available on GitHub at \url{https://github.com/eBrain4Everyone/PennySynth}.
\section*{Acknowledgment}
 This work was supported in part by the NYUAD Center for Quantum and Topological Systems (CQTS), funded by Tamkeen under the NYUAD Research Institute grant CG008, and Center for CyberSecurity (CCS), funded by Tamkeen under the NYUAD Research Institute Award G1104. This research was carried out on the High Performance Computing resources at New York University Abu Dhabi.
\bibliographystyle{IEEEtran}
\bibliography{refs.bib}

\end{document}